\documentclass[journal,a4paper]{IEEEtran}
\IEEEoverridecommandlockouts
\usepackage{cite}
\usepackage{amsmath,amssymb,amsfonts}
\usepackage{mathtools}
\usepackage{algorithmic}
\usepackage{graphicx}
\usepackage{textcomp}
\usepackage{csquotes}
\usepackage{xcolor}
\usepackage{adjustbox}
\usepackage{hyperref}
\usepackage{orcidlink}
\usepackage{multirow}
\usepackage{soul}
\usepackage{tabularx}
\usepackage{makecell}

\usepackage{subcaption}
\def\BibTeX{{\rm B\kern-.05em{\sc i\kern-.025em b}\kern-.08em
    T\kern-.1667em\lower.7ex\hbox{E}\kern-.125emX}}
\DeclareMathOperator{\argmin}{argmin}

\begin{document}

\title{Transformer-Based Robust Underwater Inertial Navigation in Prolonged Doppler Velocity Log Outages}

\author{
\IEEEauthorblockN{Zeev Yampolsky\IEEEauthorrefmark{2}\orcidlink{0009-0003-9122-7576},
% ~\IEEEmembership{Graduate Student Member,~IEEE}, 
Nadav Cohen\IEEEauthorrefmark{2}\orcidlink{0000-0002-8249-0239},
% ~\IEEEmembership{Graduate Student Member,~IEEE}, 
and Itzik Klein\orcidlink{0000-0001-7846-0654},
% ~\IEEEmembership{Senior Member,~IEEE}
}\\

\IEEEauthorblockA{\textit{The Hatter Department of Marine Technologies,} \\
\textit{Charney School of Marine Sciences,} \\
\textit{University of Haifa, Haifa, Israel}}

\thanks{\IEEEauthorrefmark{2}Zeev Yampolsky and Nadav Cohen contributed equally to this work.}
\thanks{Corresponding author: Zeev Yampolsky (zyampols@campus.haifa.ac.il).}
\thanks{Nadav Cohen (ncohe140@campus.haifa.ac.il).}
\thanks{Itzik Klein (kitzik@univ.haifa.ac.il).}

}
\maketitle

\begin{abstract}
Autonomous underwater vehicles (AUV) have a wide variety of applications in the marine domain, including exploration, surveying, and mapping. Their navigation systems rely heavily on fusing data from inertial sensors and a Doppler velocity log (DVL), typically via nonlinear filtering. The DVL estimates the AUV's velocity vector by transmitting acoustic beams to the seabed and analyzing the Doppler shift from the reflected signals. However, due to environmental challenges, DVL beams can deflect or fail in real-world settings, causing signal outages. In such cases, the AUV relies solely on inertial data, leading to accumulated navigation errors and mission terminations. To cope with these outages, we adopted ST-BeamsNet, a deep learning approach that uses inertial readings and prior DVL data to estimate AUV velocity during isolated outages. In this work, we extend ST-BeamsNet to address prolonged DVL outages and evaluate its impact within an extended Kalman filter framework. Experiments demonstrate that the proposed framework improves velocity RMSE by up to 63\% and reduces final position error by up to 95\% compared to pure inertial navigation. This is in scenarios involving up to 50 seconds of complete DVL outage.
\end{abstract}


\section{Introduction}
% New part by Zeev
\noindent
In the field of marine robotics, autonomous underwater vehicles (AUV) are utilized in a broad range of marine missions, including underwater pipeline inspection \cite{jacobi2015autonomous, zhang2022submarine} and oceanographic research \cite{paull2013auv}. As  AUVs operate as a fully autonomous platform, accurate navigation is critical for mission success. Consequently, AUVs are equipped with a wide range of navigational sensors \cite{jain2015review}. For example,  an inertial measurement unit (IMU) which include a tri-axial accelerometer (measures the specific force vector), and a tri-axial gyroscope (measures the platform’s angular velocity vector). These inertial measurements are integrated by the strapdown inertial navigation system (SINS) to compute the AUV’s navigation solution, that is, its position, velocity, and orientation \cite{titterton2004strapdown,groves2015principles}.
However, due to the inherent drift in the SINS solution, an accurate (external) aiding sensor is required to mitigate the error. In this context, the Doppler velocity log (DVL) is commonly employed for that purpose \cite{farrell2007gnss, shin2002accuracy}.\\ 
\noindent
The DVL is an acoustic sensor that transmits acoustic beams to the sea floor, which are reflected back to the DVL. Thus, by utilizing the Doppler effect, the DVL can estimate the AUV velocity vector with respect to the sea floor \cite{braginsky2020correction}.
To achieve accurate navigation through the fusion of the SINS and the DVL measurements, commonly an estimation filter such as an extended Kalman filter (EKF) is used \cite{titterton2004strapdown,chatfield1997fundamentals,farrell2008aided}.\\
In real-world scenarios,  not all beams are reflected back to the DVL due to several reasons, including an uneven sea floor, fish school blocking the beams, and extreme maneuvering by the AUV \cite{cohen2022libeamsnet,yona2021compensating}. The most extreme scenario is a complete DVL outage \cite{cohen2023set}, where all four beams are missing and the DVL cannot provide a velocity update. Thus, resorting to a pure  inertial navigation solution. \\
Lately, data-driven frameworks have been employed in the navigation domain, showing their impressive applicability to this research field \cite{cohen2024inertial}. Data-driven frameworks have even been employed for DVL-related tasks such as DVL calibration \cite{YAMPOLSKY2025104525}. Additionally, in partial DVL measurement scenarios, where only part of the beams are missing, deep learning (DL) has been used to regress the missing beams \cite{cohen2022libeamsnet,yona2021compensating}.\\
Recently, we examined Set-transformer BeamsNet (ST-BeamsNet), which employed an advanced deep-learning architecture and transformer architecture \cite{vaswani2017attention,lee2019set} to estimate all missing beams of the DVL using only three past DVL and current IMU measurements to estimate and outperform the baseline approach successfully. This work \cite{cohen2023set} was verified on data collected by Haifa's University AUV, the "Snapir," in a sea experiment, thus showcasing its robustness. This naturally raises the question: Can ST-BeamsNet be employed during prolonged DVL outages to generate surrogate DVL measurements and serve as an aiding sensor for the EKF?
To answer this question, in this work, we propose ST-AidedEKF, which employs the ST-BeamsNet deep-learning framework to regress the missing DVL measurements and provide them to the EKF as though the DVL measurements are continuously provided and no outage scenario occurred. We evaluate our proposed approach on a dataset acquired in the Mediterranean Sea and introduced in \cite{cohen2025AKIT}. We show that our proposed approach, ST-AidedEKF, was able to maintain comparable accuracy even in prolonged DVL outages scenario of up to 50 seconds. \\
\noindent
The rest of the paper is organized as follows: Section \ref{problem_form} presents the mathematical formulation of the problem, including DVL velocity modeling and the EKF used for sensor fusion. Section \ref{st_aidedekf_main_sec} describes the ST-BeamsNet architecture and details the proposed ST-AidedEKF framework that integrates predicted DVL velocities into the navigation pipeline. Section \ref{exp_res_subsec} discusses the dataset and training procedure and evaluates the proposed approach under various DVL outage scenarios. Finally, Section \ref{con} concludes the paper with key findings.

\section{Problem Formulation}\label{problem_form}
\noindent
This section presents the mathematical formulation, beginning with DVL velocity estimation procedure, followed by the EKF framework.
\subsection{DVL Velocity Updates}\label{DVL_subsec}
\noindent
The DVL is an acoustics sensor that transmits four acoustics beams to the sea floor, which are, in turn, reflected back to the DVL. Commonly, the beams are arranged in a "$\times$" configuration. Using the frequency shift, the DVL can estimate each beam velocity \cite{brokloff1994matrix}.
\begin{figure}[h]
	\centering
		\includegraphics[width=\columnwidth]{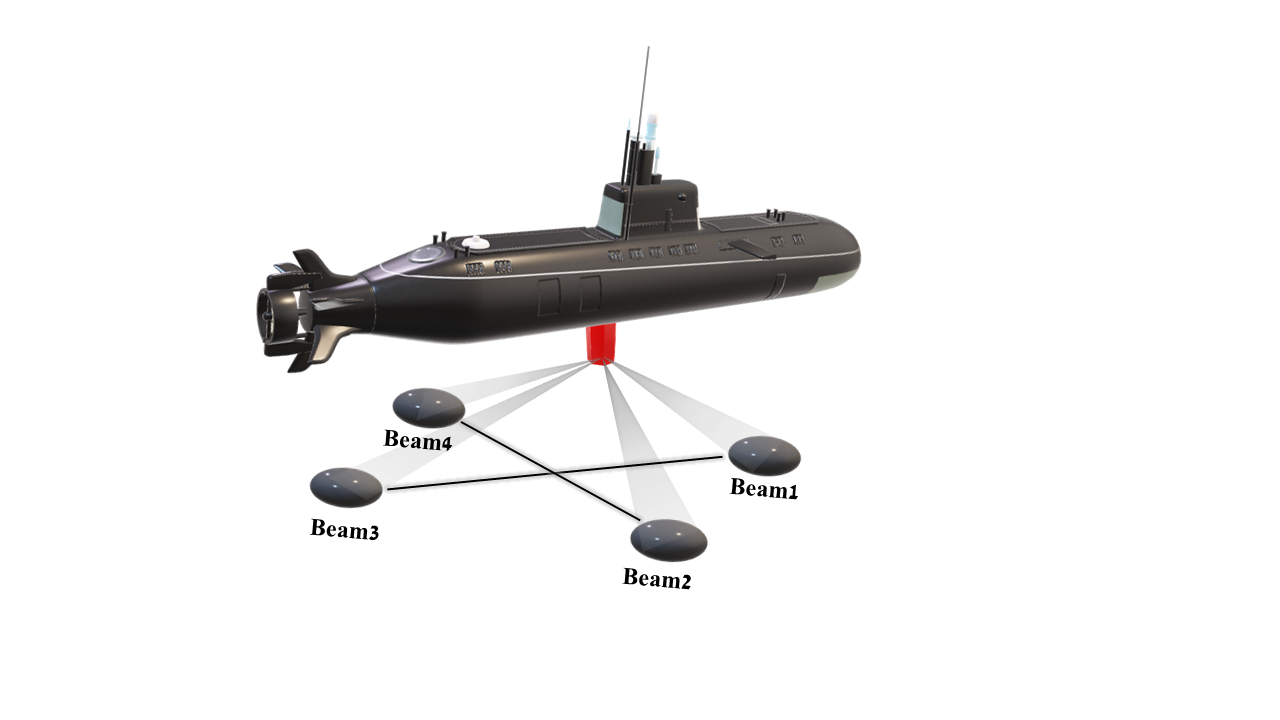}
	  \caption{A DVL transmits acoustic beams in \enquote*{$\times$} shape configuration.}\label{fig1}
\end{figure}
Each beam direction can expressed by  \cite{cohen2022beamsnet} \cite{braginsky2020correction}: 
\begin{equation}\label{eqn:1}
    \centering
        \boldsymbol{beam}_{\dot{\imath}}=
        \begin{bmatrix} 
        \cos{\psi_{\dot{\imath}}}\sin{\theta}\quad
        \sin{\psi_{\dot{\imath}}}\sin{\theta}\quad
        \cos{\theta}
    \end{bmatrix}_{1\times3}
\end{equation} 
where $\theta$ is a constant pitch angle and is the same for all beams, $\psi$ is the yaw angle of each beam, and $\dot{\imath}$ represents the beam's number. The yaw angle is defined as:
\begin{equation}\label{eqn:2}
    \centering
        \psi_{\dot{\imath}}=(\dot{\imath}-1)\cdot\frac{\pi}{2}+\frac{\pi}{4}\;[rad]\;,\; \dot{\imath}=1,2,3,4
\end{equation}
By stacking all beams projections, $\textbf{b}_{i}$, the matrix $\mathrm{H}$ is constructed:
\begin{equation}\label{h_stacking}
    \centering
    \mathbf{A}=
    \begin{bmatrix} 
    \boldsymbol{beam}_{1}\\\boldsymbol{beam}_{2}\\\boldsymbol{beam}_{3}\\\boldsymbol{beam}_{4}\\
    \end{bmatrix}_{4\times3}
\end{equation}
Assuming the DVL frame and the body frame coincide, the beam velocity vector can be represented as follows \cite{liu2018ins}:
\begin{equation}\label{beams_vel}
    \centering
    \boldsymbol{\upsilon}_{beam}=\mathbf{A}\boldsymbol{v}_{b}^{b}
\end{equation}
Commonly, to solve \eqref{beams_vel}, the following least squares (LS) problem is defined:
\begin{equation}\label{eqn:5}
    \centering
        {\boldsymbol{v}}_{b}^{d}=
        \underset{\boldsymbol{v}_{b}^{d}}{\argmin}{\mid\mid\boldsymbol{y}-\mathbf{A}\boldsymbol{v}_{b}^{d} \mid\mid}^{2}
\end{equation} 
where $\boldsymbol{y}$ is the measured beams velocity vector. The common error model applied over the beams' velocity measurements is given by \cite{cohen2025AKIT}:
\begin{equation}\label{beams_error_model}
    \centering
        \tilde{\boldsymbol{y}} = [\mathbf{A} \boldsymbol{v}_{\mathrm{b}}^{d}(1+\boldsymbol{s}_{\mathrm{DVL}})]+\boldsymbol{b}_{\mathrm{DVL}}+\boldsymbol{\sigma}_{\mathrm{DVL}}
\end{equation}
where $\tilde{\boldsymbol{y}}$ is the measured beams velocity, $\boldsymbol{s}_{\mathrm{DVL}}$ is the beams scale factor, $\boldsymbol{b}_{\mathrm{DVL}}$ is the beams bias vector, and $\boldsymbol{\sigma}_{\mathrm{DVL}}$ is the Gaussian zero mean white noise.
To estimate the velocity measured by the DVL in the body frame, \eqref{beams_error_model} is substituted into \eqref{eqn:5}, and the least squares (LS) solution is obtained using the pseudoinverse, as follows \cite{braginsky2020correction}:
\begin{equation}\label{eqn:6}
    \centering
        \tilde{\boldsymbol{v}}_{b}^{d}=(\mathbf{A}^{T}\mathbf{A})^{-1}\mathbf{A}^{T}\tilde{\boldsymbol{y}}
\end{equation} 
%----------------
\subsection{EKF Formulation for INS/DVL Fusion}\label{ekf_subsec}
\noindent
This subsection presents the formulation of an error-state EKF. Three primary coordinate systems are used throughout the derivation, each indicated via superscripts. The body frame, $()^b$, is fixed to the vehicle and centered at its center of mass. It is assumed that the sensitive axes of the inertial sensor coincide with the body frame.
The navigation frame, $()^n$, corresponds to a local-level frame aligned with the North-East-Down (NED) convention. Lastly, the DVL frame, denoted $()^{DVL}$, is sensor-specific and defined according to the manufacturer’s reference. The transformation from the DVL frame to the body frame is assumed to be fixed and known \cite{farrell2008aided}.
\\ \noindent
The EKF uses an error-state formulation, where the deviation between the estimated and true states is modeled as an additive perturbation. Specifically, let the error-state vector be denoted $\boldsymbol{\delta x} \in \mathbb{R}^{n\times1}$, capturing the discrepancy between the true state $\boldsymbol{x}^{t}$ and the estimated state $\boldsymbol{x}^{e}$ as:
\begin{equation}
\label{eqn:errormodel}
\boldsymbol{x}^{t} = \boldsymbol{x}^{e} - \boldsymbol{\delta x}
\end{equation}
In the case of INS/DVL fusion, twelve error states are considered, leading to the following error-state vector:
\begin{equation}
\label{eqn:dx}
\boldsymbol{\delta x} = \begin{bmatrix} \boldsymbol{(\delta v^{n})}^{T} & \boldsymbol{(\epsilon^{n})}^{T} & \boldsymbol{(\delta b_{a})}^{T} & \boldsymbol{(\delta b_{g})}^{T} \end{bmatrix}^{T} \in \mathbb{R}^{12\times1}
\end{equation}
where $\boldsymbol{\delta v^{n}}$ is the velocity error in the navigation frame, $\boldsymbol{\epsilon^{n}}$ denotes the attitude error, and $\boldsymbol{\delta b_{a}}$ and $\boldsymbol{\delta b_{g}}$ represent biases in the accelerometer and gyroscope measurements, respectively \cite{rogers2003applied}.
The dynamics of the error state evolve according to the linearized system:
\begin{equation}
\label{eqn:f}
\boldsymbol{\delta \dot{x}} = \mathbf{F} \boldsymbol{\delta x} + \mathbf{G} \boldsymbol{w}
\end{equation}
with $\boldsymbol{w} \in \mathbb{R}^{12\times1}$ representing the process noise vector. The system matrix $\mathbf{F} \in \mathbb{R}^{12\times12}$ governs the propagation of the error state, while $\mathbf{G} \in \mathbb{R}^{12\times12}$ modulates the impact of process noise. This noise is assumed to be zero-mean Gaussian, composed of several independent components:
\begin{equation}
\label{eqn:noise}
\boldsymbol{w} = \begin{bmatrix} \boldsymbol{w_{a}}^{T} & \boldsymbol{w_{g}}^{T} & \boldsymbol{w_{a_b}}^{T} & \boldsymbol{w_{g_b}}^{T} \end{bmatrix}^{T}
\end{equation}
where $\boldsymbol{w_{a}}$ and $\boldsymbol{w_{g}}$ are additive white noise terms for the accelerometer and gyroscope, respectively, and $\boldsymbol{w_{a_b}}, \boldsymbol{w_{g_b}}$ are modeled as random walks representing bias drift.
The system matrix $\mathbf{F}$ and noise shaping matrix $\mathbf{G}$ can be found in the literature, for example, in \cite{groves2015principles,farrell2008aided}.
\\ \noindent
The EKF operates in two main stages: prediction and correction. In the prediction step, the error-state vector is initialized to zero:
\begin{equation}
\label{eqn:apriorizero}
\delta \boldsymbol{x}^{-} = 0
\end{equation}
The covariance matrix $\mathbf{P}^{-}_k$ is then propagated forward using the system transition matrix $\mathbf{\Phi}_{k-1}$ and the process noise covariance $\mathbf{Q}_{k-1}$:
\begin{equation}
\label{eqn:pred}
\mathbf{P}^{-}_{k} = \mathbf{\Phi}_{k-1} \mathbf{P}^{+}_{k-1} \mathbf{\Phi}_{k-1}^{T} + \mathbf{Q}_{k-1}
\end{equation}
The transition matrix $\mathbf{\Phi}_{k-1}$ is obtained via a truncated Taylor series expansion:
\begin{equation}
\label{eqn:transition}
\mathbf{\Phi}_{k-1} = \sum_{r=0}^{\infty} \frac{(\mathbf{F}_{k-1} \tau_s)^r}{r!}
\end{equation}
The discrete-time process noise covariance $\mathbf{Q}_{k-1}$ is approximated using a mid-point rule \cite{gelb1974applied}:
\begin{equation}
\label{eqn:Q_dis_MB}
\mathbf{Q}_{k-1} = \frac{1}{2} \left( \mathbf{\Phi}_{k-1} \mathbf{G}_{k-1} \mathbf{Q} \mathbf{G}_{k-1}^{T} + \mathbf{G}_{k-1} \mathbf{Q} \mathbf{G}_{k-1}^{T} \mathbf{\Phi}_{k-1}^{T} \right) \Delta t
\end{equation}
\\ \noindent
In the update phase, measurements are incorporated using the Kalman gain:
\begin{equation}
\label{gain}
\mathbf{K}_{k} = \mathbf{P}^{-}_k \mathbf{H}_k^{T} \left( \mathbf{H}_k \mathbf{P}^{-}_k \mathbf{H}_k^{T} + \mathbf{R}_k \right)^{-1}
\end{equation}
\begin{equation}
\label{postriori}
\mathbf{P}^{+}_k = \left( \mathbf{I} - \mathbf{K}_k \mathbf{H}_k \right) \mathbf{P}^{-}_k
\end{equation}
\begin{equation}
\label{dz}
\boldsymbol{\delta x}^{+}_k = \mathbf{K}_k \boldsymbol{\delta z}_k
\end{equation}
where $\mathbf{H}_k$ is the observation model, $\mathbf{R}_k$ is the measurement noise covariance, and $\boldsymbol{\delta z}_k$ is the innovation vector. For DVL updates the measurement matrix is defined by: 
\begin{equation}
    \centering
    \mathbf{H}_{k} = \begin{bmatrix}
        \mathbf{C}_{n}^{b} & -\mathbf{C}_{n}^{b}\boldsymbol{v}^{n}[\times] & \mathbf{0}_{3\times3} & \mathbf{0}_{3\times3}
    \end{bmatrix}
\end{equation}
where $\mathbf{C}_{n}^{b}$ is the transformation matrix from the navigation from to the body frame, $\boldsymbol{v}^{n}$ is the AUV velocity in the body frame, $[\times]$ is the skew operation, $\mathbf{0}_{3\times3}$ is a $3\times3$ zero valued matrix.
%To compute $\mathbf{H}_k$, one must examine the DVL sensor configuration as presented in the previous subsection \ref{DVL_subsec}, and its derivation can be found in \cite{cohen2024seamless}.
\section{Proposed approach}\label{st_aidedekf_main_sec}
\begin{figure}
    \centering
    \includegraphics[width=0.95\columnwidth]{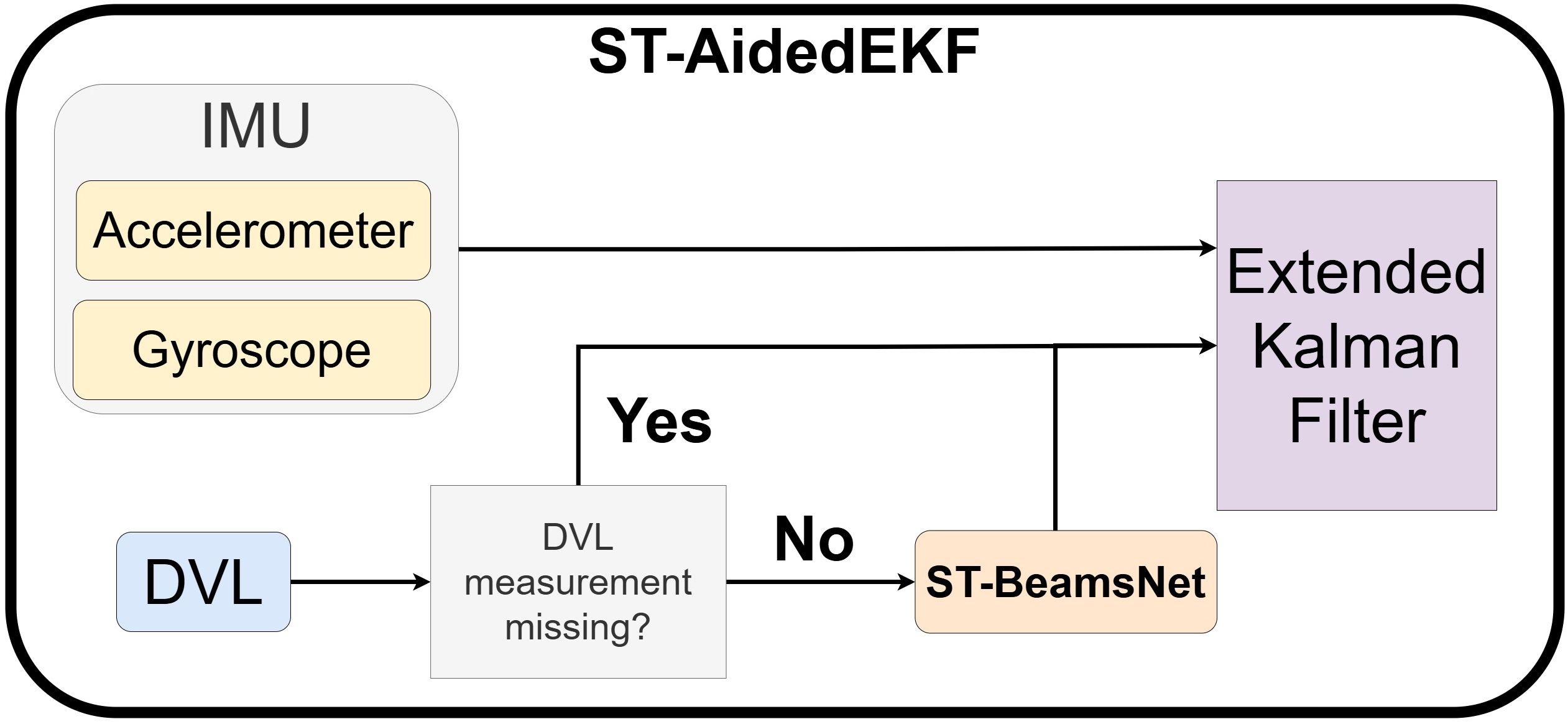}
    \caption{Block diagram of the ST-AidedEKF framework.}
    \label{fig:st_prop_diag}
\end{figure}

\noindent
Our proposed approach, set transformer aided EKF (ST-AidedEKF) forecasts the DVL-based velocity vector during DVL outages using the set transformer BeamsNet (ST-BeamsNet) approch \cite{cohen2023set}. Once regressed, the velocity vector is introduced into the EKF, as though no DVL outage occurred. Figure \ref{fig:st_prop_diag} presents a general overview and flow of our proposed approach ST-AidedEKF.  The rest of this section first introduces the mathematical formulation for ST-BeamsNet, and later the describes in detail the overall ST-AidedEKF framework.
% --------------------------------
\subsection{ST-BeamsNet Network Architecture}\label{st_arch_blocks}
\begin{figure*}
    \centering
    \includegraphics[width=0.98\textwidth]{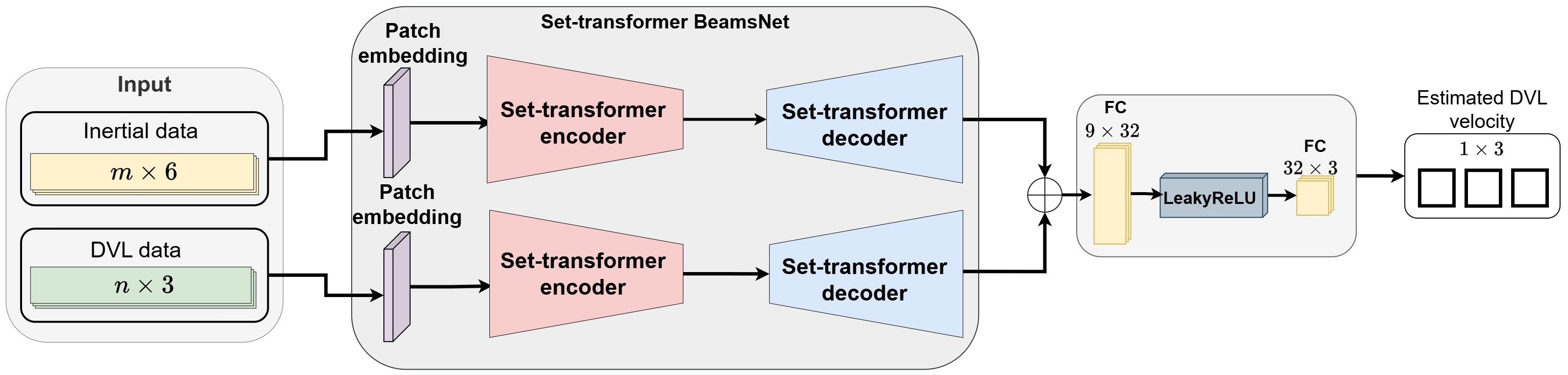}
    \caption{Block diagram illustrating the 
    ST-BeamsNet architecture which process the DVL and IMU measurements to regress the missing DVL velocity vector due to DVL outage. The input to ST-BeamsNet are the $n$ past DVL measurements, and $m$ past IMU measurements.}
    \label{fig:st_general_arch}
\end{figure*}

\noindent
The ST-BeamsNet model is constructed using modular components from the Set Transformer \cite{lee2019set} framework, designed to process sequential sensor data and extract meaningful temporal and spatial patterns. This section outlines the core learning blocks that make up the architecture. Figure \ref{fig:st_general_arch} presents an overview of the ST-BeamsNet architecture.
\\ \noindent
The first component is the \textbf{Patch Embedding} block. It applies a one dimensional convolution to segment the input time series into local patches, while extracting useful features from the input and project them into a latent space. Three hyperparameters control this process:
\begin{itemize}
    \item Patch width: denoted as $\alpha$, it is the convolution kernel size.
    \item Kernel stride: denoted as $\beta$, and defines the stride between each two computations of the convolution kernel and defines the shift between patches. 
    \item Patch size: denoted as $\gamma$, which defines the patch size, and determines the patch receptive field.
    \item Latent dimension: denoted as $D$, this hyperparameter defines the dimension of the latent vector, which encodes the data for the following block, the set transformer encoder.
\end{itemize}
%ZY i put it in comment because i think in list is better in this case
% the  convolution kernel size $\alpha$, which determines the patch width, the stride $\beta$, which sets the shift between patches, and also is the stride of the convolution kernel over the input, and the patch size $\gamma$, which controls the receptive field. The number of output filters from this operation defines the latent dimensionality and is denoted as $D$.\\
\noindent
Following the patch embedding, the model includes a sequence of \textbf{Set Attention Block}s (SABs). In a SAB there are several multi-head self-attention heads \cite{vaswani2017attention}, which each self-attention head learns complex relationships and dependencies between all input elements, which are the output of the PE block. The number of SABs used in the encoder is given by $b$, and each attention block uses $h$ attention heads.
\\ \noindent
Each SAB contains a feed-forward network with a dimensional expansion controlled by the feed-forward expansion factor $FFE$. This feed-forward network expansion section allows the model to increase its representational capacity before projecting the output back to the latent dimension $d$, which is typically set equal to $D$.
\\ \noindent
Once the latent information is passed through the SAB, the network uses \textbf{Pooling by Multihead Attention} (PMA) to reduce the high dimensional set of encoded features into a compact form. This pooling mechanism introduces $k$ trainable vectors that learn to extract even more information from the sequence and refine the extracted information. These vectors serve as queries in an attention mechanism, enabling the model to extract information and features better, thus promoting learning. This describes the PE and set-transformer encoder block as presented in the ST-BeamsNet architecture in Figure \ref{fig:st_general_arch}.\\
\noindent The information and extracted features which are the output of the ST encoder block, are passed next through a lightweight decoder composed of additional attention and feed-forward layers, which refine the information and map it to the output latent space, typically a predicted velocity vector, and in our case a $\mathbb{R}^{3}$ velocity vector.
\\ \noindent
A more in-depth formulation of the set transformer can be found in \cite{lee2019set,cohen2023set}. All hyperparameters mentioned here—$\alpha$, $\beta$, $\gamma$, $D$, $d$, $h$, $FFE$, $b$, and $k$ are summarized in Table \ref{st_beamnet_hyp_tbl}. They define the capacity and flexibility of the network across its various components.
\begin{table*}[!ht]\caption{Set-transformer BeamsNet hyperparameters}\label{st_beamnet_hyp_tbl}
\begin{adjustbox}{width=\textwidth}
\begin{tabular}{|c|c|c|c|c|c|c|c|c|}
\hline
Description & \begin{tabular}[c]{@{}c@{}}kernel\\ size\end{tabular} & \begin{tabular}[c]{@{}c@{}}kernel\\ stride\end{tabular} & \begin{tabular}[c]{@{}c@{}}patch\\ size\end{tabular} & \begin{tabular}[c]{@{}c@{}}PE latent\\ dim\end{tabular} & \begin{tabular}[c]{@{}c@{}}SAB\\ number\end{tabular} & \begin{tabular}[c]{@{}c@{}}Att head\\ num\end{tabular} & FF Exp & \begin{tabular}[c]{@{}c@{}}Pool\\ vectors\end{tabular} \\ \hline
Notation    & $\alpha$                                 & $\beta$ & $\gamma$ & $D$ & $b$ & $h$& $FFE$  & $k$                \\ \hline
Value       & 200                                                   & 100                                                     & 1                                                  & 128                                                     & 16                                                  & 2                                                    & 256                                                      & 3                                                    \\ \hline
\end{tabular}
\end{adjustbox}
\end{table*}
\subsection{Set Transformer Aided EKF}
\noindent
Our proposed approach, ST-AidedEKF intervenes when a DVL outage scenario occurs. In normal operating conditions, the velocity updates from the DVL are passed to the EKF as external information updates. In a scenario of DVL outage, where the DVL cannot estimate the DVL velocity, the ST-BeamsNet is utilized to forecast the AUV velocity vector based on two quantities: the first is the available past DVL measurements, which correspond to $n$ past measurements, and the second is the available past IMU measurements, which correspond to $m$ past measurements.
The estimated velocity by the ST-BeamsNet is passed now to the EKF and treated as a velocity update, as though no DVL outage scenario occurred, enabling continuous velocity updates, assuring high navigation accuracy. \\
\noindent 
The input to ST-AidedEKF are the available past $n$ DVL measurements, and the available past $m$ inertial measurement from the accelerometer and gyroscope. Notice that the sampling rates of the IMU and DVL differ, thus affecting the exact numerical value of $n$ and $m$, thus they will be fully discussed in Section \ref{datasets_subsec}.\\
\noindent 
As described in Section \ref{st_arch_blocks}, both the inertial and DVL measurements are fed to the network as input, processed simultaneously by two PE blocks, and later by two separate ST encoder and encoder blocks. The two separate outputs are concatenated together and are passed through two fully connected layers \cite{sze2017efficient}, the first is followed by a Droupout layer \cite{srivastava2014dropout} to promote generalization and then by a hyperbolic tangent (Tanh) function \cite{sharma2017activation}. The output of the FC block is a $\mathbb{R}^{3}$ vector which is the estimated $n+1^{th}$ AUV velocity vector as would be provided by a DVL with all available beams.
% Our proposed approach leverages data from the IMU, along with with $n$ past DVL velocity measurements. The ST-BeamsNet is designed to extract meaningful features from each sensor modality using a patch embedding layer. These features are then analyzed for interdependencies through a multi-head attention mechanism. Subsequently, a learnable pooling aggregation is applied to capture the most informative features and relationships. Finally, fully connected layers are used to fuse the extracted information and regress the AUV's velocity vector. Specifically, the model utilizes \( n+1 \) seconds of inertial sensor data sampled at 100~Hz, corresponding to the current time step and the preceding \( n \) seconds. In parallel, it incorporates \( n \) past DVL velocity measurements sampled at 1~Hz.
% \\ \noindent  
% The information flow in the proposed network proceeds as follows: the input data first passes through a patch embedding layer, followed by $b$ blocks consisting of encoder–decoder modules. The outputs from the decoder blocks, corresponding to the different sensor modalities, are then fused and processed through two fully connected layers with an ReLU activation function in between. As the network is designed to predict the $(n+1)$-th velocity vector, the mean squared error (MSE) is employed as the loss function. A block diagram of the proposed ST-AidedEKF architecture is shown in Figure~\ref{fig:st_prop_diag}.
% \label{dataset_exp_sec}
% --------------------------------------
\section{Experimental Results}\label{exp_res_subsec}
\noindent To verify our proposed approach, ST-AidedEKF, real-world experimental data was recorded and ST-AidedEKF was trained and verified using the collected data. In this section we describe the dataset, trainning process, and present the results.
% -----------------------------------------------
\subsection{Datasets}\label{datasets_subsec}
\begin{figure}[!ht]
	\centering
		\includegraphics[width=\columnwidth]{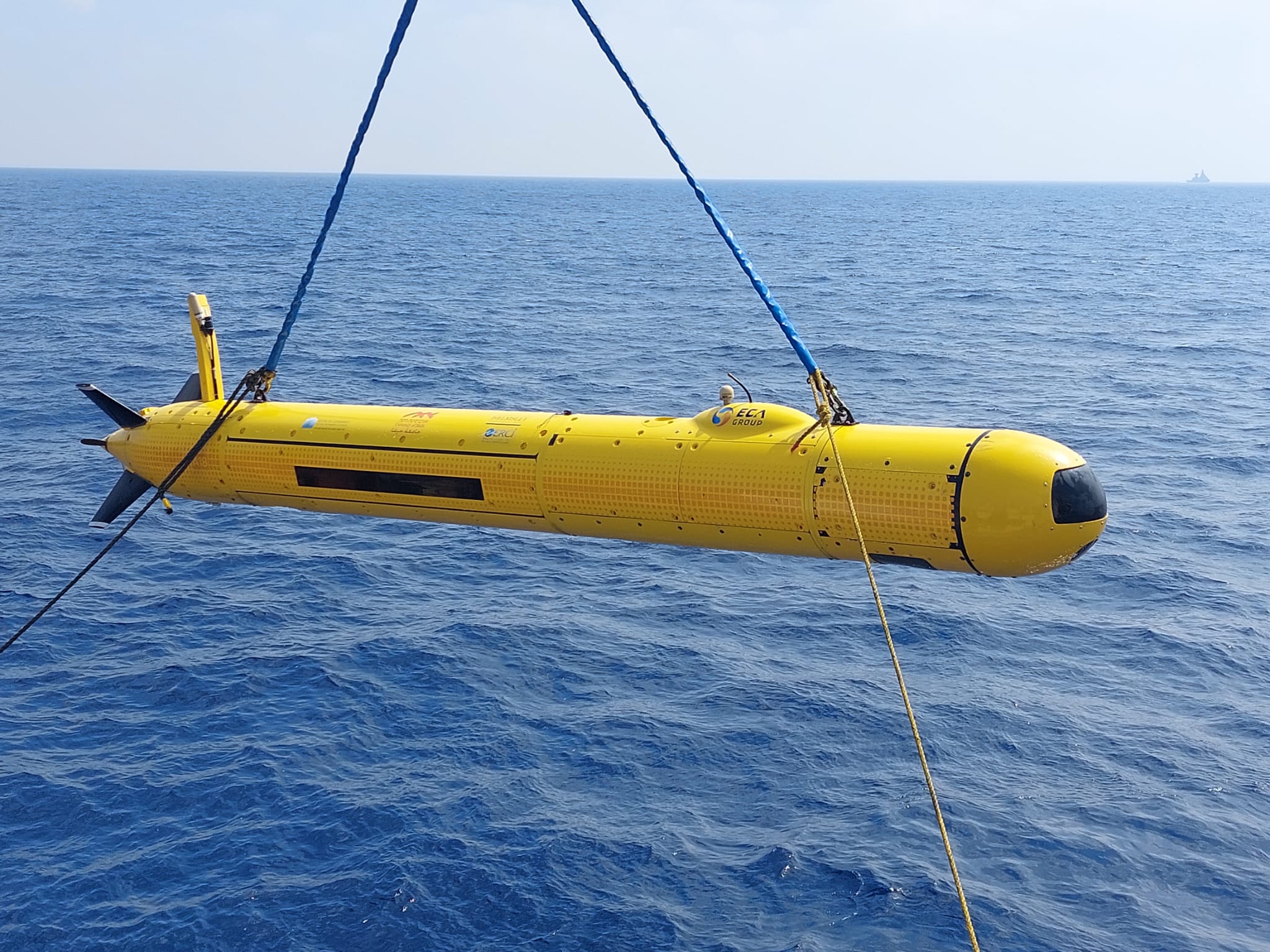}
	  \caption{Snapir AUV being lowered into the Mediterranean sea before an experiment.}\label{fig:snapir}
\end{figure}

\noindent
To evaluate our proposed approach, ST-AidedEKF, data were collected using the University of Haifa's AUV, "Snapir", Figure~\ref{fig:snapir}. Snapir is a 5.5-meter-long AUV with a 0.5-meter diameter. It is equipped with a high-end DVL, the Navigator DVL~\cite{Teledyne}, which samples at 1~Hz, and a high-end IMU, the iXblue Phins Subsea IMU~\cite{iXblue}, which samples at 100~Hz. The dataset is publicly available at~\cite{cohen2025AKIT}. The onboard navigation computer on Snapir provides the navigation solution based, and the DVL and IMU data is supplied.
\\ \noindent
During the sea experiments, thirteen missions ,M1-M13, were recorded, totaling 87 minutes of data. Eleven missions ,M1-M11, were used for training ST-AidedEKF, while the remaining two missions ,M12 and M13, were reserved for evaluation of the proposed approach.
% -----------------------
\subsection{Training and Evaluation}\label{train_and_eval_sec}
\noindent
% -----------------------
To train and evaluate ST-AidedEKF, all thirteen missions ,M1-M13, were passed through a noising pipeline that added zero-mean Gaussian white noise to the inertial measurements, resulting in the measured values which are fed to ST-AidedEKF $\tilde{\boldsymbol{f}}^b$ and $\tilde{\boldsymbol{\omega}}^{b}$. The values of the velocity random walk and angular random walk which were added to the accelerometer and gyroscope as the addative noise is presented in Table~\ref{tbl:imu_noise}.
\begin{table}[!ht]
\centering
\caption{The velocity random walk (VRW) and angular random walk (ARW) values of the zero-mean Gaussian white noise added to the raw accelerometer and gyroscope measurements, respectively.
}
\label{tbl:imu_noise}
\begin{tabularx}{\columnwidth}{|>{\centering\arraybackslash}X|
                                >{\centering\arraybackslash}X|}
\hline
      VRW & ARW \\ \hline
57 [$\mu g /\sqrt{Hz}$] & 0.018 [$^\circ /\sqrt{Hz}$] \\ \hline
\end{tabularx}
\end{table}
% IK remove column 1 from the table - ZY: Done
To train ST-AidedEKF, every four seconds of data were processed by removing every fourth DVL measurement while retaining the corresponding four seconds of past inertial measurements. Due to the sampling rate differences between the DVL and IMU, this procedure resulted in $n=3$ past DVL measurements and $m=400$ IMU measurements. 
Notice that in Figure \ref{fig:st_general_arch} we do not refer to the sampling rate differences, and present the input in units of number of samples, hence $n\times3$ and $m\times6$. \\
\noindent 
Each DVL input is of size $3 \times 3$, where the first dimension corresponds to the three past measurements and the second to the three velocity components, this is since DVL velocity is expressed in the body frame. Each IMU input is of size $(6 \times 400)^T$, where the six channels represent the stacked accelerometer $\tilde{\boldsymbol{f}}^b$ and gyroscope $\tilde{\boldsymbol{\omega}}^b$ vectors, and the 400 samples represent four seconds of data sampled at 100~Hz. Note that the last second of IMU samples corresponds in time to the missing DVL beams due to the DVL outage scenario, which are the velocity vector that ST-BeamsNet estimates.\\
\noindent TNext, the DVL and IMU measurements were concatenated to construct a dataset of 4,356 data points. This dataset was shuffled and randomly divided into training and testing subsets using a 75\%:25\% ratio.
During training, ST-AidedEKF predicted the missing fourth DVL measurement due to DVL outage, based on $n=3$ past DVL measurements and $m=400$
%IK 400 or 600 as you wrote before??? - ZY: 400, i probably had a type
of IMU measurements. The predicted fourth DVL measurement was then compared to the ground truth to compute the mean squared error (MSE) loss, which was used to update the network’s weights and biases. Both the training and testing sets were divided into batches of size 128, and ST-AidedEKF was trained for 500 epochs on a computer equipped with an NVIDIA RTX 4090 GPU and 16~GB of RAM. An MSE score was computed at the end of each epoch, and the model with the lowest RMSE on the validation set was saved for use in the evaluation missions, M12 and M13.
\\ \noindent
To evaluate the performance of ST-AidedEKF, the following evaluation parameters were defined:
\begin{itemize}
    \item \textbf{Outage duration}: The length of time during which a complete DVL outage occurs. Three outage durations were evaluated: 30, 40, and 50 seconds. The outage duration is denoted as \( t_{\text{duration}} \).
    
    \item \textbf{Outage start time}: The time at which the DVL outage begins, denoted as \( t_{\text{init}} \). It is measured in seconds.
\end{itemize}
The motivation for this evaluation stems from the inherent uncertainty in real-time maneuvering, where the timing and duration of a complete DVL outage cannot be controlled during real-world missions and experiments. To simulate this variability, five different outage start times were randomly selected. At each start time, \( t_{\text{init}} \), the three previously defined outage durations were evaluated by removing \( t_{\text{duration}} \) seconds of DVL measurements following each \( t_{\text{init}} \). This process was performed for both evaluation missions. The result is a comprehensive evaluation procedure that captures the variability and unpredictability of complete DVL outage scenarios. Moreover, the selected outage durations reflect prolonged outage conditions, allowing for a robust assessment of ST-AidedEKF under challenging operational circumstances.
\\ \noindent
For each evaluation mission, and for each combination of outage duration and start time, ST-AidedEKF was applied to estimate the missing DVL velocity measurements. Once these velocities were estimated, the EKF was able to compute the AUV’s velocity and orientation, as it now had access to a full set of velcoity aiding measurements. In contrast, the baseline approach, referred to as PureINS, relied solely on the inertial measurements to estimate the AUV’s velocity and orientation. In this approach, from the moment of outage initiation at \( t_{\text{init}} \), the DVL measurements were unavailable for \( t_{\text{duration}} \) seconds, and no DVL aiding was provided to the EKF during this period. Therefore, for a duration of $t_{duration}$ the EKF solution relied solely on the IMU measurements.
% ----------------------
\subsection{Results}\label{Res_subsec}
\begin{figure}[!ht]
    \centering
    \includegraphics[width=0.95\linewidth]{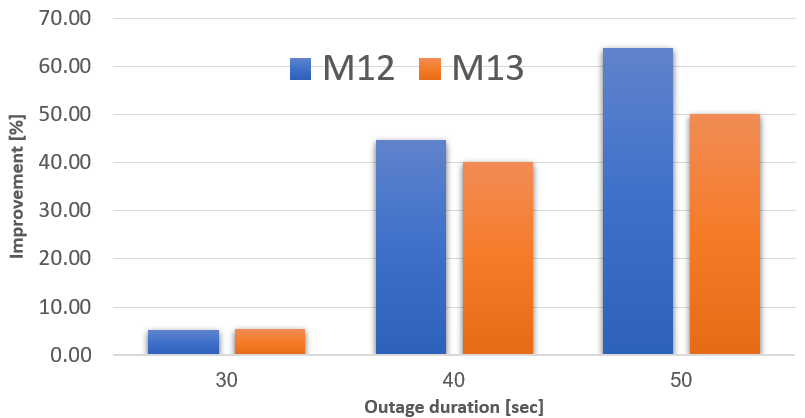}
    \caption{Improvement $[\%]$ of the average velocity RMSE of ST-AidedEKF over the baseline, for each outage duration. The blue columns represent the results of mission M12, and the orange columns represent the results of mission M13.}
    \label{fig:vel_rmse_improv}
\end{figure}

\begin{figure}[h!]
        \centering
        \begin{subfigure}{0.475\linewidth}
            \includegraphics[width=\linewidth]{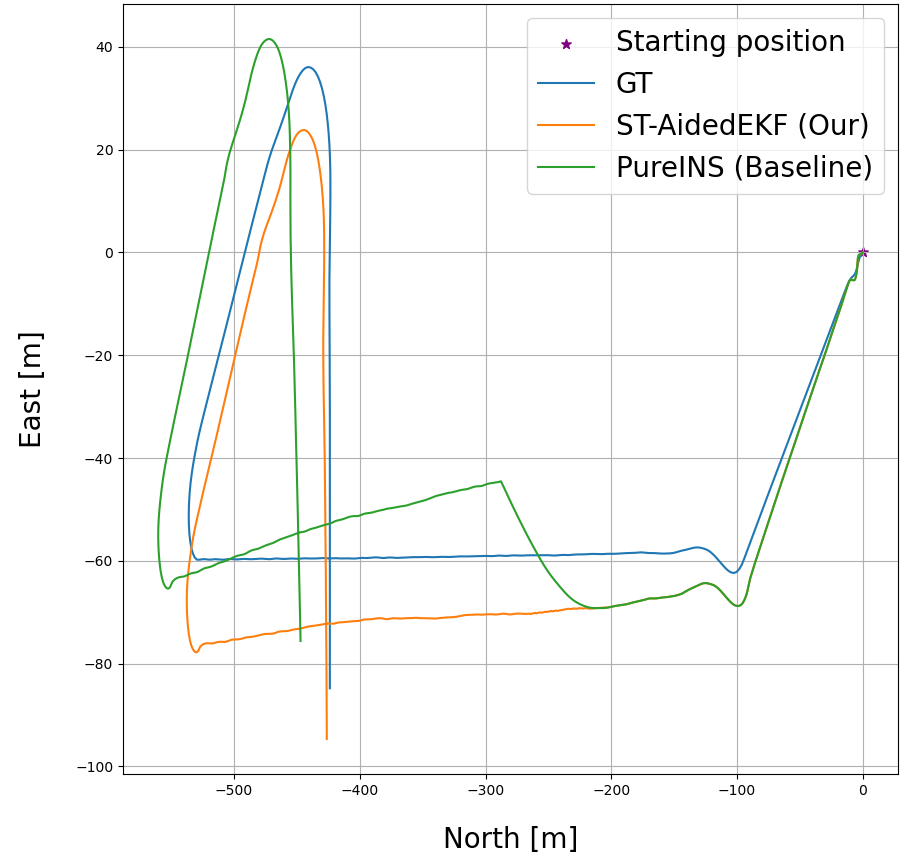}
            \caption{M12 at $t_{duration} = 20$}  
            \label{subfig:M12_30_1}
        \end{subfigure}
        \hfill
        \begin{subfigure}[b]{0.475\linewidth}  
            \includegraphics[width=\linewidth]{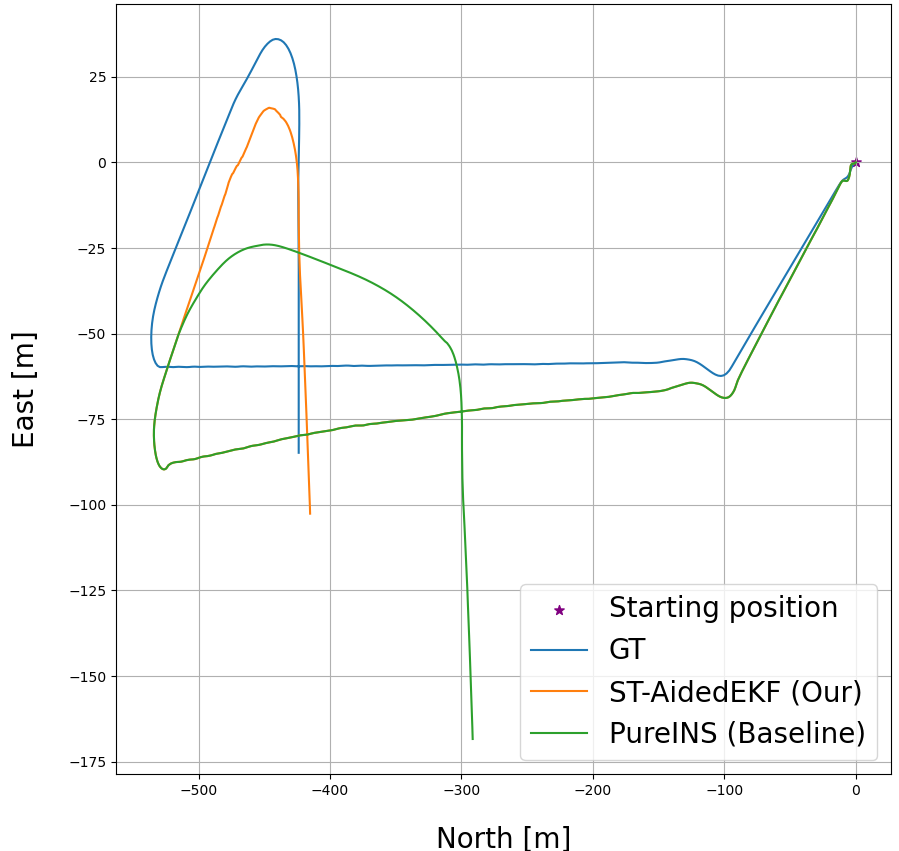}
            \caption{M12 at $t_{duration} = 50$}  
            \label{subfig:M12_50_3}
        \end{subfigure}
        % \medskip
        \vskip\baselineskip
        \begin{subfigure}{0.475\linewidth}   
            \includegraphics[width=\linewidth]{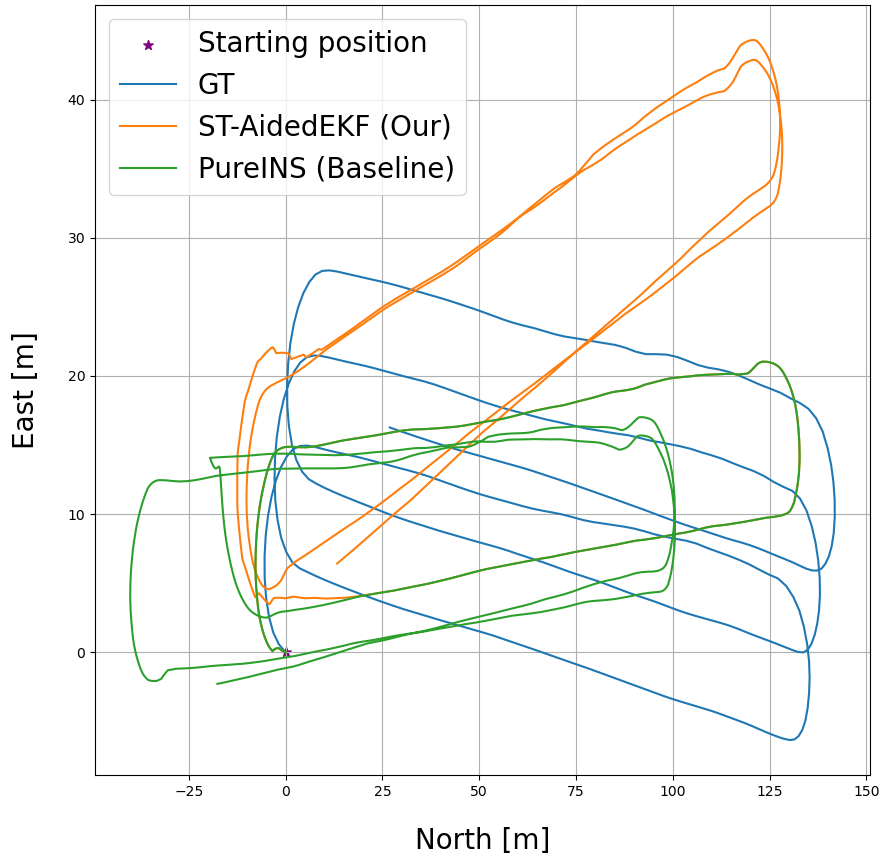}
            \caption{M13 at $t_{duration} = 30$}
            \label{subfig:M13_30_2}
        \end{subfigure}
        \hfill
        \begin{subfigure}[b]{0.475\linewidth}   
            \includegraphics[width=\linewidth]{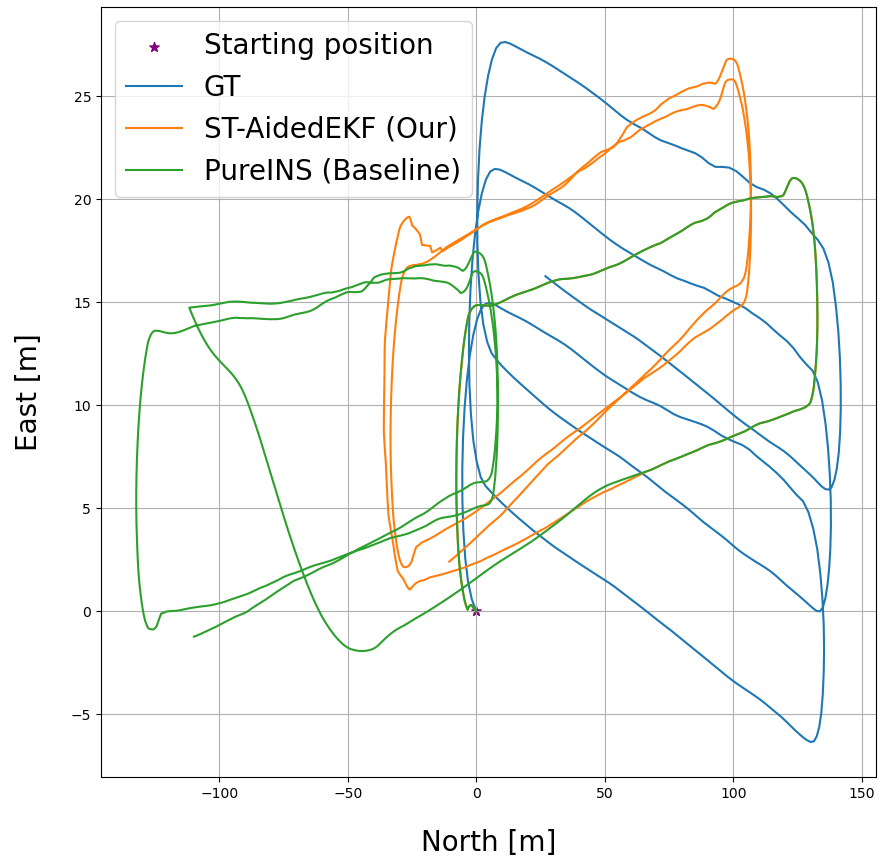}
            \caption{M13 at $t_{duration} = 50$}
            \label{subfig:M13_50_5}
        \end{subfigure}
    \caption{
Estimated positions in the North and East plane for ST-AidedEKF (our) in organe, the baseline (PureINS) in green, the ground truth (GT) in blue, and the purple star is the starting point. Figure \ref{subfig:M12_30_1} and \ref{subfig:M12_50_3} show the position estimates for mission M12 under 30 second and 50 second DVL outage durations, respectively. Figures \ref{subfig:M13_30_2} and \ref{subfig:M13_50_5} present the same for mission M13.}\label{fig:2d_plots_of_12_13}
\end{figure}

\begin{figure}
\centering
\begin{subfigure}{\columnwidth}
  \centering
  \includegraphics[width=.99\linewidth]{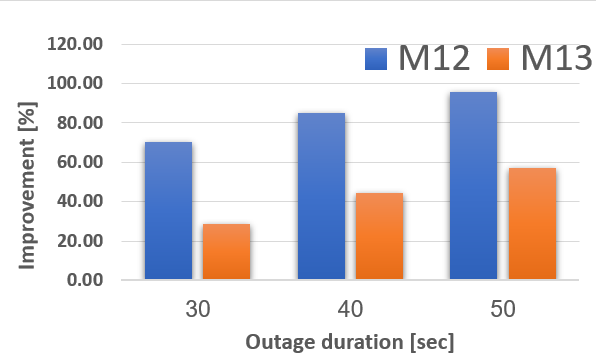}
  \caption{Final position error meter}
    \label{subfig:final_pos_error}
\end{subfigure}%
\\
\begin{subfigure}{\columnwidth}
  \centering
  \includegraphics[width=.99\linewidth]{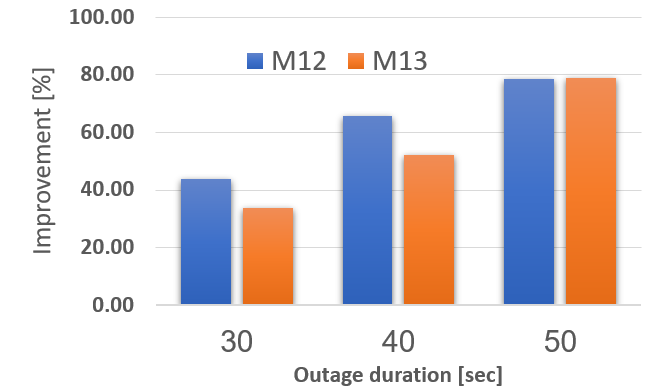}
  \caption{Position RMSE [meter]}
    \label{subfig:rmse_pos_err}
\end{subfigure}
\caption{
Average improvement of ST-AidedEKF in different outage durations for the two evaluation missions M12 and M13. Figure~\subref{subfig:final_pos_error} shows the improvement in final point position error, while Figure~\subref{subfig:rmse_pos_err} presents the improvement in position RMSE.
}
\label{fig:pos_fin_err_and_rmse}
\end{figure}

\noindent
We evaluated ST-AidedEKF performance based on the following two metrics:
\begin{enumerate}
    \item  The root mean squared error (RMSE)
    \begin{equation}\label{eqn:18}
    \centering
    \mathrm{RMSE}(\boldsymbol{x}_{i} , \hat{\boldsymbol{x}_{i}}) = \sqrt{\frac{\sum_{i=1}^{N} \sum_{j = X,Y,Z}(\boldsymbol{x}_{i,j} - \hat{\boldsymbol{x}}_{i,j})^{2}} {N}}
\end{equation}
\item  Absolute final position error:
    \begin{equation}\label{eqn:19}
    \centering
    \mathrm{AFPE}(x_{n} , \hat{x}_{n}) = \frac{1}{3}\sum_{j=X,Y,Z}|x_{j} - \hat{x}_{j}|
    \end{equation}
    \end{enumerate}
%         MFPE(\boldsymbol{x}_{\dot\imath},\hat{\boldsymbol{x}}_{\dot\imath})=\frac{\sum_{\dot\imath=1}^{N}|\boldsymbol{x}_{\dot\imath}-\hat{\boldsymbol{x}}_{\dot\imath}|}{N}
% \end{equation}
%     \item  Mean Absolute Error (MAE)
%     \begin{equation}\label{eqn:19}
%     \centering
%         MAE(\boldsymbol{x}_{\dot\imath},\hat{\boldsymbol{x}}_{\dot\imath})=\frac{\sum_{\dot\imath=1}^{N}|\boldsymbol{x}_{\dot\imath}-\hat{\boldsymbol{x}}_{\dot\imath}|}{N}
% \end{equation}
where $\boldsymbol{x}_{\dot\imath}$ is the GT velocity or position vector $\boldsymbol{v}_{GT}^{b}$ at time step $i$, $\hat{\boldsymbol{x}}_{\dot\imath}$ is the estimated vector by ST-AidedEKF, $N$ is the number of time samples, and $x_{n}$ is the position vector at the last time step $n$. Note that a perfect RMSE \eqref{eqn:18} score is $0$ \cite{armaghani2021comparative} and a perfect final position error \eqref{eqn:19} is also $0$, thus meaning there is no position difference between our ST-AidedEKF and the GT.\\ \noindent
To evaluate the performance of ST-AidedEKF, we compared the velocity estimates estimated by the EKF when aided by ST-BeamsNets predicted DVL measurements to those produced by a baseline EKF that relied solely on inertial measurements (i.e., PureINS) during the DVL outage period. Table~\ref{tbl:velocity_RMSE} presents the average velocity RMSE computed across the five randomly selected outage start times.
%IK change the table's caption. dont use misses selected - ZY - Did it, but maybe recheck if its OK. - NC
\begin{table}[!h]
\centering
\caption{Average velocity RMSE $[m/s]$ across five different outage start times, shown for each outage duration and both evaluation trajectories.}\label{tbl:velocity_RMSE}
\begin{adjustbox}{width = 0.95\columnwidth}
\begin{tabular}{|c|c|c|c|}
\hline
\begin{tabular}[c]{@{}c@{}}Mission\\ \end{tabular} & \begin{tabular}[c]{@{}c@{}}Outage\\ duration$[s]$\end{tabular} & ST-AidedEKF $[m/s]$& PureINS $[m/s]$\\ \hline
\multirow{3}{*}{M12}                                    & 30                                                      & 0.95        & 1.68    \\ \cline{2-4} 
                                                        & 40                                                      & 1.06        & 3.03    \\ \cline{2-4} 
                                                        & 50                                                      & 1.13        & 4.57    \\ \hline
\multirow{3}{*}{M13}                                    & 30                                                      & 0.88        & 1.01    \\ \cline{2-4} 
                                                        & 40                                                      & 0.91        & 1.6     \\ \cline{2-4} 
                                                        & 50                                                      & 1.03        & 2.25    \\ \hline
\end{tabular}
\end{adjustbox}
\end{table}
From Table~\ref{tbl:velocity_RMSE}, it is evident that our proposed approach, ST-AidedEKF, consistently outperforms the baseline approach across all evaluated outage durations. Accordingly, Figure~\ref{fig:vel_rmse_improv} illustrates the average improvement in velocity RMSE achieved by ST-AidedEKF over the baseline approach for both M12 and M13 across each evaluated outage duration.
From Figure~\ref{fig:vel_rmse_improv}, it is evident that the improvement achieved by ST-AidedEKF increases with the duration of the DVL outage. To further evaluate the effectiveness of ST-AidedEKF, we integrated the velocity estimates produced by the EKF. These velocity estimates are expressed in the NED coordinate frame. 
Figure~\ref{fig:2d_plots_of_12_13} presents the resulting two dimensional trajectories for both evaluation missions, M12 and M13, as estimated by the velocities by ST-AidedEKF, the baseline PureINS and the GT.
From Figure~\ref{fig:2d_plots_of_12_13}, it is evident that ST-AidedEKF enables reasonably accurate position estimation during DVL outages. However, visual inspection alone does not clearly indicate whether ST-AidedEKF or the baseline PureINS performs better in terms of position accuracy. Therefore, two quantitative metrics were used to evaluate the position error:  the first, is the final point position error (AFPE) \eqref{eqn:19}, and the second the RMSE \eqref{eqn:18} of the position over the entire outage duration. The complete comparison of these position errors is provided in Table~\ref{tbl:pos_error_comp}.
%IK add to column 1 Mission and Outage Duration $[s]$. - ZY: - Done
\begin{table}[h!]
\centering
\caption{Final position error and position RMSE were estimated by the integration of ST-AidedEKF velocity estimation and the baseline approaches.}
\label{tbl:pos_error_comp}
\resizebox{\columnwidth}{!}{%
\begin{tabular}{|cc|cc|cc|}
\hline
\multicolumn{2}{|c|}{\multirow{2}{*}{\makecell{Mission and \\ Outage Duration $[s]$}}} 
& \multicolumn{2}{c|}{Final position error [m]} & \multicolumn{2}{c|}{Position RMSE [m]} \\ \cline{3-6} 
\multicolumn{2}{|c|}{}                          & \multicolumn{1}{c|}{ST-AidedEKF} & Baseline & \multicolumn{1}{c|}{ST-AidedEKF} & Baseline \\ \hline
\multicolumn{1}{|c|}{\multirow{3}{*}{M12}} & 30 & \multicolumn{1}{c|}{5.59}        & 24.73    & \multicolumn{1}{c|}{16.31}       & 36.53    \\ \cline{2-6} 
\multicolumn{1}{|c|}{}                     & 40 & \multicolumn{1}{c|}{5.66}        & 53.64    & \multicolumn{1}{c|}{17.03}       & 69.44    \\ \cline{2-6} 
\multicolumn{1}{|c|}{}                     & 50 & \multicolumn{1}{c|}{6.25}        & 7.11     & \multicolumn{1}{c|}{17.4}        & 117.62   \\ \hline
\multicolumn{1}{|c|}{\multirow{3}{*}{M13}} & 30 & \multicolumn{1}{c|}{12.28}       & 18.07    & \multicolumn{1}{c|}{19.74}       & 29.77    \\ \cline{2-6} 
\multicolumn{1}{|c|}{}                     & 40 & \multicolumn{1}{c|}{13.88}       & 26.56    & \multicolumn{1}{c|}{21.98}       & 50.45    \\ \cline{2-6} 
\multicolumn{1}{|c|}{}                     & 50 & \multicolumn{1}{c|}{15.37}       & 42.33    & \multicolumn{1}{c|}{23.9}        & 11.09    \\ \hline
\end{tabular}%
}
\end{table}
According to Table~\ref{tbl:pos_error_comp}, ST-AidedEKF achieves lower final position error and position RMSE compared to the baseline approach for both M12 and M13 trajectories. Furthermore, it maintains its superior accuracy even under prolonged DVL outage scenarios. To better illustrate the improvement, Figure~\ref{fig:pos_fin_err_and_rmse} presents a bar plot showing the percentage improvement of ST-AidedEKF over the baseline in both error metrics.
Figure~\ref{fig:pos_fin_err_and_rmse} demonstrates the substantial improvement in position error achieved by the proposed ST-AidedEKF architecture over the PureINS baseline. In the most prolonged DVL outage scenario, with a duration of 50 seconds, ST-AidedEKF achieves an impressive reduction of approximately 95\% in final point position error for mission M12. Additionally, it shows an improvement of around 80\% in overall position RMSE for both M12 and M13 under the same outage condition.
% -------------------------------
\section{Conclusions}\label{con}
\noindent
In this work, we introduced {ST-AidedEKF}, a novel framework designed to maintain continuous and accurate DVL updates even during complete DVL outage scenarios. This is achieved by leveraging ST-BeamsNet to predict the missing beam measurements, which are then used as velocity updates in the EKF. To train and evaluate the proposed approach, data from thirteen real-world sea missions totaling 87 minutes were collected using the University of Haifa's AUV, Snapir.
\\ \noindent
ST-AidedEKF was evaluated on two unseen missions from the dataset and demonstrated strong performance in estimating the AUV velocity vector using inertial and past DVL measurements. In the most prolonged outage scenario of 50 seconds, it achieved up to a 63\% improvement in velocity RMSE compared to the baseline PureINS approach. Furthermore, it exhibited excellent results in position estimation, with up to an 80\% improvement in position RMSE.
\\ \noindent
To summarize, this work presents a novel framework that enables continuous velocity updates for the navigation filter even in the presence of prolonged complete DVL outages.
\section*{Acknowledgments}
\noindent
N.C. and Z.Y. are supported by the Maurice Hatter Foundation and the University of Haifa presidential scholarship for students on a direct Ph.D. track. 
\bibliographystyle{IEEEtran}
\bibliography{refs}

\end{document}